\documentstyle[12pt,amsmath,amsfonts,theorem]{article}

\theoremstyle{break}

\newtheorem{definition}{Definition}

\newtheorem{theorem}{Theorem}

\newcommand{\set}[1]{{\mathbb{#1}}}

\begin{document}
\title{ \vspace{-1.7 cm}
Complexity analysis for algorithmically simple strings}
\author{Andrei N. Soklakov\footnote{e-mail: a.soklakov@rhul.ac.uk}\\
\\
{\it Department of Mathematics} \\
{\it Royal Holloway, University of London}\\
{\it Egham, Surrey TW20 0EX, United Kingdom}}

\date{25 February 2002}

\maketitle

\begin{abstract}
\vspace{-9mm}
Given a reference computer, Kolmogorov complexity is
a well defined function on all binary strings. 
In the standard approach, however, only
the asymptotic properties of such functions are considered
because they do not depend on the reference computer.
We argue that this approach can be more useful if it is refined
to include an important practical case of simple binary strings.
Kolmogorov complexity calculus may be developed
for this case if we restrict the class of available reference computers.
The interesting problem is to define a class of computers
which is restricted in a {\it natural} way modeling the
real-life situation where only a limited class of computers
is physically available to us. We give an example of what such a natural
restriction might look like mathematically, and show that under such
restrictions some error terms, even logarithmic in complexity, can
disappear from the standard complexity calculus.

\end{abstract}

\section{Introduction}

The asymptotic nature of Kolmogorov complexity 
calculus renders it significantly less useful in practical applications
such as inference by the minimum description length (MDL)
principle~\cite{Rissanen_1978}.
In the classical MDL approach~\cite{Rissanen_1997}
this problem is solved by replacing
Kolmogorov complexity with a phenomenological
complexity measure just before performing the actual inference.
Such a measure can be chosen to suit a particular application,
whereas the general form of the MDL constructions can be
considered as a consequence of the asymptotic properties
of Kolmogorov complexity (consult section 5.5 in Ref.~\cite{LiVitanyi}).
Here we propose a different
approach. We argue that Kolmogorov complexity can become
more practical if we restrict the class of reference computers.

Computer science is not the only field which can benefit
from the proposed research. There is a growing
interest in using Kolmogorov complexity as a fundamental
{\it physical} concept. This includes applications in
thermodynamics~\cite{Bennett_1982,Bennett_1987,Zurek_1989}%
\footnote{consult~\cite{LiVitanyi} for further references.}, theory of
chaos~\cite{Brudno_1978,Brudno_1982,Ford_1983,SchackCaves_1992}%
\footnote{
consult~\cite{LiVitanyi} for further references.},
physics of 
computation~(consult~\cite{LiVitanyi} and references therein),
and many other areas of modern theoretical 
physics~\cite{Dzhunushaliev,SoklakovSchack,Soklakov00}.
It is however very difficult to use Kolmogorov complexity
in any concrete physical setting, or indeed, in any concrete
application. For that we need a much more detailed
calculus that can be applied to particular cases of reference computers.
The main aim of this article is to stimulate
further research in developing such a {\it practical} complexity calculus.

This article is organized as follows.
In section~\ref{Basic} we review some basic definitions.
In section~\ref{Main} we present the main conceptual arguments
of the paper. In section~\ref{Example} we give an example of how
one can build a restricted class of computers in a ``natural'' way.
Considering one of the central equalities of the standard complexity
calculus we give an illustration of how the error terms may be reduced.

\section{Basic definitions} \label{Basic}

Let
$\set{X}=\{\Lambda,0,1,00,01,10,11,000,\dots\}$
be the set of
finite binary strings where $\Lambda$ is the string of length 0.
A set of strings $\set{Y}\subset \set{X}$ with the property that no string in
$\set{Y}$ is a prefix of another is called an instantaneous code.
 A prefix computer is a partial recursive
function
$C: \set{Y}\times \set{X}\to \set{X}$.
For each $p\in \set{Y}$ (program string) and for
each $d\in \set{X}$ (data string) the output of the computation is either
undefined or given by $C(p,d)\in \set{X}$.
Kolmogorov complexity
of a string $\alpha$ given a data string $d$ relative to a computer
$C$ is defined as the length
$K_C(\alpha|d)$ of the shortest program that
makes $C$ compute $\alpha$ given data~$d$:
\begin{equation}
K_C(\alpha|d)\equiv\min_{p}\{ |p|\; {\big{|}}\;C(p,d)=\alpha\}\,,
\end{equation}
where $|p|$ denotes the length of the program $p$ (in bits).

Since this complexity measure depends strongly on the
reference computer, it is important to find an optimal computer $U$ such
that the complexity of any string relative to $U$ is not much higher that
the complexity of the same string relative to any other computer $C$.
Mathematically, a computer $U$ is called optimal if
\begin{equation}
\forall C\ \ \exists\kappa_C\ \mbox{such that } \forall \alpha,d:\ \  K_U(\alpha|d)\leq K_C(\alpha|d)
+\kappa_C\,,
\end{equation}
 where $\kappa_C$ is a constant depending
on $C$ (and $U$) but not on $\alpha$ or $d$. 
It turns out that the set of prefix computers contains such a $U$ and,
moreover, it can be constructed so that any prefix computer
can be simulated by $U$: for further details consult~\cite{LiVitanyi}.
Such a $U$ is called a universal prefix computer and its choice is not unique.
Using some particular universal prefix computer $U$ as a reference,
the conditional Kolmogorov complexity of $\alpha$
given $\beta$ is defined as $K_U(\alpha|\beta)$.

The above definitions are generalized for the case
of many strings as follows. We choose and fix a particular recursive bijection
$B: \set{X}\times \set{X}\to \set{X}$ for use throughout the rest of this paper.
 Let $\{\alpha^i\}_{i=1}^{n}$
be a set of $n$ strings $\alpha^i\in \set{X}$.
For $2\leq k\leq n\;$ we define
${\langle\alpha^1,\alpha^2,\dots,\alpha^k\rangle}\equiv
B({\langle\alpha^1,\dots,\alpha^{k-1}\rangle},\alpha^k)$,
and ${\langle\alpha^1\rangle}\equiv\alpha^1$.
We can now define $K_U(\alpha^1,\dots,\alpha^n| \beta^1,\dots,\beta^k)
\equiv K_U({\langle \alpha^1,\dots,\alpha^n\rangle }|{\langle\beta^1,
\dots,\beta^k\rangle})$.

For any two universal prefix computers $U_1$ and $U_2$ we have, by
definition, $|K_{U_1}(\alpha|\beta) -K_{U_2}(\alpha|\beta)|
\leq \kappa(U_1,U_2)$
where $\kappa(U_1,U_2)$ is a constant that depends only on $U_1$
and $U_2$ and not on $\alpha$ or $\beta$. Most of the research on
Kolmogorov complexity is focused on the asymptotic case of
nearly random long strings, when $\kappa(U_1,U_2)$ can
be neglected in comparison to the value of the complexity.
In such cases, Kolmogorov
complexity becomes an asymptotically absolute measure of the
complexity of individual strings. For this reason,
many fundamental properties of Kolmogorov complexity are established
up to an error term which is asymptotically small compared to the
complexity of strings involved. For instance,  the standard
analysis of the prefix Kolmogorov
complexity~(\cite{LiVitanyi}, Section 3.9.2)
gives
\begin{equation}            \label{ErrorDelta}
K_U(\alpha,\gamma|\beta)=K_U(\alpha|\gamma,\beta)
                                                       +K_U(\gamma|\beta)+\Delta\,,
\end{equation}
where $\Delta$ is an error term which grows logarithmically
with the complexity of the considered strings. This is an example
of an asymptotic property that all Kolmogorov measures of complexity
have irrespective of the choice of the reference computer.
Of course, it is important to know that all Kolmogorov measures
of complexity share many of their asymptotic properties.
For any given reference computer, however,
Kolmogorov complexity is a well defined function on all binary strings.
Even from a purely mathematical viewpoint it is interesting
to study the properties of such functions beyond the asymptotics.
As for the applied viewpoint, consider, by analogy, mathematical
analysis. This theory would be much less useful if we studied
only asymptotic properties of functions.

\section{Main arguments}\label{Main}

Without significant knowledge about the reference computer,
Kolmogorov complexity can be considered only up to an additive
error term $O(1)$.
Error terms even as small as $O(1)$ make it impossible
to use Occam's razor to discriminate between simple
hypotheses. The importance of this problem becomes
apparent once we recognize that the domain of simple hypotheses
is absolutely crucial in our every-day life as well as in fundamental
science. Indeed, it is often the case that, after extensive analysis,
the greatest scientific discoveries can be expressed in a form so simple
that they are readily understood by even school children.

Humans can relatively easily discriminate between different hypotheses
even when the Kolmogorov complexities involved are rather small.
This gives them an enormous advantage over the present-day
theoretical models. A good example is Kepler's theory of planetary motion.
In what was
a major breakthrough in theoretical astronomy at the time,
Kepler introduced elliptical orbits as a better alternative to the complicated
Copernican planetary model of superimposed epicycles.
At the level of accuracy provided by Brahe's experiments, the original
Copernican model had to be refined by introducing additional
epicycles: the Keplerian theory appeared to be simpler
and therefore better by Occam's razor. This apparently obvious
fact cannot be established using the standard formalism
of Kolmogorov complexity: whereas Kepler's theory can be simpler
relative to some type of computers, the Copernican model can be
simpler relative to some other type of reference computers.

Much simpler examples can be found in tests that are
designed by humans to test their own intelligence.
A typical problem in such tests is to find the
next element in a sequence of symbols. For example,
if the first four elements of a sequence are 1,2,3,4 
an intelligent person is supposed to see the simplest
pattern and predict 5 as the next element of the sequence.
As in the previous example, all humans would agree that
predicting 5 would correspond to the choice of the simplest
hypothesis, whereas the standard formalism of Kolmogorov
complexity cannot be used to justify this.
It seems entirely plausible that
the ultimate theory of artificial intelligence and,
in particular, inductive
inference, can achieve human-like results only if the
building blocks of the theory, such as Kolmogorov complexity,
are made sensitive to small variations in the complexity of hypothesis.

The $O(1)$ ambiguity in the classical definition of Kolmogorov complexity
and the error terms like $\Delta$ in Eq.~(\ref{ErrorDelta})
is the price we pay for having an unrestricted class of reference computers.
Every human perceives complexity with respect to their own
built-in reference computer -- the brain.
As in the case of abstract reference computers,
human brains are not identical. However, they are similar enough to
allow for a sharper discrimination between individual theories
on the basis of their complexity. This suggests that further progress
in applications of Kolmogorov complexity to the theory of induction
can be made possible if we find a natural way
of restricting the class of reference computers.

We see from this discussion that some restrictions on the
class of reference computers are needed. 
It is desirable, however, to have a complexity
theory which would be as general as possible. As a compromise,
we can try to group all possible reference computers into restricted
classes. Although we may want to study all such classes,
we can argue that due to biological, technological, and
other limitations only one class of reference computers is
physically available to us.
A definition of this realistic class of reference computers would
be the crucial link between the abstract theory of
Kolmogorov complexity and the practical theories of induction and
computer learning.

What kind of restriction of the class of reference computers can be
seen as natural? It appears natural to assume that given some particular
level of technology one can build more powerful computers only at the
expense of a more complex internal design. 
In section~\ref{Example} we use this observation
to construct an example of a ``natural''
restriction of the class of reference computers. 
Roughly speaking, this restriction entails
the requirement that switching to a more complex reference
computer should always be accompanied by an
equivalent reduction of program lengths.
Using some particular universal computer $U$
as a reference, we define the complexity of a computer $W_s$
from the set $\{W_i\}$ given data $d$ as $K_U(s|d)$.
We then construct a particular set of computers $\{W_i\}$
such that the sum of the complexity of a computer and the length
of a program for it is the same for all 
equivalent\footnote{two programs $p_1$ and $p_2$ for computers
$C_1$ and $C_2$ are called equivalent iff $C_1(p_1|d)=C_2(p_2|d)$.}
programs and for all
computers in the set $\{W_i\}$
(consult section~\ref{Example} for details).
This gives us a tradeoff between computer complexity and
program lengths similar to what one would expect in the
real world where we face various practical limitations.
Together with the original reference computer $U$
computers $\{W_i\}$ form a ``naturally'' restricted class.
 It is natural to define a
computer $W$ which is universal for this class by setting $W(p,\langle
s,d\rangle)=W_s(p,d)$, where $U$ is included by defining
$W_\Lambda\equiv U$.
Using any such $W$ as a reference we can see that, in principle,
even error terms logarithmic in complexity can
be removed from the standard complexity calculus. In particular,
we prove that for any triple of simple strings $\alpha,\beta,\gamma$,
we have
 \begin{equation}                       \label{Kw}
K_W(\alpha,\gamma|\langle\Lambda,\beta\rangle)=
     K_W(\alpha|\gamma,\beta)
   +K_W(\gamma|\langle\Lambda,\beta\rangle)+{\mbox{\rm const}}\,,
\end{equation}
where the constant depends only on the reference machine $W$
(not on $\alpha$, $\beta$ or~$\gamma$). Apart from subtleties
associated with the operation of combining strings into pairs, this
is analogous to Eq.~(\ref{ErrorDelta}) with the important difference
that the error term is replaced by a constant.

In the standard complexity calculus the above equation holds only up to
an error term which grows logarithmically with the complexity
of the considered strings. As we explained earlier, this is unacceptable
if we want to analyze the complexity of simple strings. 
The error terms
are especially troublesome if we want to use the complexity calculus
as a part of inductive inference based on the MDL principle.
In such cases we are interested in the {\it position}
of the minimum rather than on the approximated value of complexity.
The error term can significantly shift the position of the minimum
even when mistakes on the value of complexity are minor. This can
introduce uncontrollable mistakes in the inference results.
In our case, however, equation~(\ref{Kw}) is exact in the sense
that the constant does not influence the position of critical points
so it can be safely ignored in applications such as induction by the
MDL principle.

\section{Example} \label{Example}

As we explained in section~\ref{Main}, a natural restriction of the class
of reference computers can make Kolmogorov complexity more
useful in applications such as inference and computer learning.
In this section we consider one possible way of making such a restriction.
We show that, in the important case of simple strings,
the proposed restriction effectively removes the
error term in Eq.~(\ref{ErrorDelta}),
which has important applications in physics~\cite{Soklakov00}.

\begin{definition}
Fix $\delta\in\set{N}$. A set of strings $\set{S}_\delta\subseteq \set{X}$
is called $\delta$-simple iff for any two strings $\alpha,\gamma \in \set{S}_\delta$
we have
\begin{equation}
        |\alpha|<\delta\,,\ \ \ |\gamma|<\delta\,, 
        {\rm\ \ \  and\ \ \ }|\langle\alpha,\gamma\rangle|<\delta\,,
\end{equation}
where $|\cdot|$ denotes the string length.
\end{definition}

Following Chaitin \cite{Chaitin75}, consider a list of infinitely many
requirements ${\langle r_k,l_k(d)\rangle}$ $(k=0,1,2,\dots)$ for the
construction of a computer. Each requirement
${\langle r_k,l_k(d)\rangle}$ requests that a program
of length $l_k(d)$ be assigned to the result $r_k$ if the computer is given
data $d$.  The requirements are said to satisfy the Kraft inequality
if $\sum_{k}2^{-l_k(d)}\leq 1$: for such requirements there exists an
instantaneous code characterized by the set of string lengths $\{l_k(d)\}$.
A computer $C$ is said to satisfy the requirements if there are precisely as
many programs $p$ of length $l(d)$ such that $C(p,d)=r$ as there are pairs
${\langle r,l(d)\rangle}$ in the list of requirements.

Fix a universal computer $U$ which can be constructed from an effectively 
given list of requirements (consult~\cite{Chaitin75}, Theorem 3.2).
Consider the set of all programs $\{p_k\}$ for $U$
such that the output of computation $U(p_k,d)$ is defined.
Since $B$ is a bijection, we can write $U(p_k,d)={\langle r_k,s_k\rangle}$,
where $r_k$ and $s_k$ are strings from $\set{X}$.
Moreover, because $U$ is a universal computer, any pair of strings
${\langle \alpha,\gamma\rangle}$ can be generated this way.
In what follows we consider only those $p_k$ for which $s_k\neq\Lambda$.
For every fixed $s$ from
the set $\{s_k\}$ we construct a list of requirements
\begin{equation}                                                                         \label{requirements}
  {\langle r_k,|p_k|-K_U(s|d)+\kappa^s_d\rangle }\,,\ k=1,2,\dots
\end{equation}
where $|p_k|$ is the length of the program $p_k$, and $\kappa^s_d$ is some
constant.
It was shown~(\cite{Chaitin75}, Theorem 3.8)
that the constant $\kappa^s_d$ can be chosen large enough
such that these requirements satisfy the Kraft inequality.
Fix any $\delta\in\set{N}$, and consider
a sublist of requirements~(\ref{requirements}):
\begin{equation}                                                                  
{\langle r_k,|p_k|-K_U(s|d)+\kappa^s_d\rangle}
\ \ \ r_k,\, d\in \set{S}_\delta\,,
\end{equation}
where $\set{S}_\delta$ is the set of $\delta$-simple strings.
For any $s\in\set{S}_\delta$, we can find 
$\kappa\equiv\max\{\kappa^s_d|\,s,d\in \set{S}_\delta\}$,
then choose $\kappa^s_d=\kappa$, and construct a new list
of requirements
\begin{equation}                                                                      \label{requirements2}
{\langle r_k,|p_k|-K_U(s|d)+\kappa\rangle}
\ \ \ r_k,\, d\in \set{S}_\delta\,.
\end{equation}
For any fixed $s\in\set{S}_\delta$ these requirements satisfy
the Kraft inequality
by construction. Furthermore, since  $\set{S}_\delta$ is finite and
$B$ is recursive these requirements can be effectively given.
This means that for any $s\in\set{S}_\delta$ there
is a computer $W_s$ that satisfies these requirements:
consult (\cite{Chaitin75}, Theorem 3.2) for further details.

For each value of $s\in\set{S}_\delta\setminus\{\Lambda\}$ we use
(\ref{requirements2}) to construct
one $W_s$. We define $W_\Lambda=U$, and form the set 
$\set{W}_U\equiv\{W_s |\,s\in \set{S}_\delta\}$.
This set contains the original computer $U$ as a somewhat special
element. Having the computer $U$ at our disposal, it would take at least
$K_U(s|d)$ bits to specify any other $W_s$ from the set $\set{W}_U$
given data $d$. We can now see that requirements~(\ref{requirements2})
are designed
in such a way that more complex computers, i.e. larger $K_U(s|d)$,
will have shorter programs,
$l_k(d)= |p_k|-K_U(s|d)+\kappa$.
This is exactly the property that we wanted to use as a
natural restriction that defines a realistic class of computers.

In what follows we restrict our attention
to the set $\set{W}_U$. We define a computer $W$
which is universal for the set $\set{W}_U$, i.e. which is designed
to simulate any computer $W_s\in \set{W}_U$:
\begin{equation}
  W(p,{\langle s,d\rangle})\equiv W_s(p,d)\,.
\end{equation}

\begin{theorem}
For any $\alpha,d \in \set{S}_\delta$, and for any
$\gamma \in \set{S}_\delta\setminus \{\Lambda\}$, we have
\begin{equation}                                                                                  \label{KwKu}
K_{W}(\alpha|\gamma,d)
   =K_W(\alpha,\gamma|\langle\Lambda, d\rangle)
      -K_W(\gamma|\langle\Lambda, d\rangle) + \kappa\,.
\end{equation}
\end{theorem}

{\bf Proof}\\
Consider the program
$\tilde{p}_k$ which causes $W_s\in\set{W}_U$
to produce the result $r_k\in\set{S}_\delta$
given data~$d$
\begin{equation}                                                                                          \label{r}
 W_s(\tilde{p}_k,d)=r_k\,.
\end{equation}
By definition of $W_s$, the length of $\tilde{p}_k$ satisfies the
requirement
\begin{equation}                                                                                     \label{pAbs}
\forall s\in\set{S}_\delta\setminus \{\Lambda\} {\rm\ \, and\ }
 \forall d\in\set{S}_\delta: \ \ \ 
|\tilde{p}_k|=|p_k|-K_U(s|d)+\kappa\,,
\end{equation}
where $p_k$ is the program for $U$ such that
\begin{equation}                                                                                               \label{rs}
U(p_k,d)={\langle r_k,s_k\rangle }\,,\ \ s_k\neq\Lambda\,.
\end{equation}
We define the set $\set{K}\equiv\{ i |\, U(p_{i},d)={\langle r_k,s_k\rangle}\}$,
which can contain more than one element since some
of the pairs $\{{\langle r_k,s_k\rangle}\}$ can coincide.
 From the construction of
$W_s$ we note that requirements
(\ref{requirements2}) associate exactly one program $\tilde{p}_k$
with the corresponding program $p_k$. In other words there is
a one-to-one correspondence between programs
$\tilde{p}_k$ and $p_k$ (which is given explicitly by the index $k$).
This means that the set
$\set{K}$ coincides with the set $\tilde{\set{K}}
\equiv\{ i |\, W_s(\tilde{p}_{i},d)=r_k\} $.
Since $U$, $d$ and $s$ are fixed, and using the identity
$\set{K}=\tilde{\set{K}}$, we have from Eq.~(\ref{pAbs})
\begin{equation}                                                                    \label{ShortestLengths}
\min_{k\in\tilde{\set{K}}}|\tilde{p}_k|
 =\min_{k\in{\set{K}}}|p_k|-K_U(s|d)+\kappa\,,
\ \ s\in \set{S}_\delta\setminus \{\Lambda\}\,.
\end{equation}
By definition of $W$ we have
\begin{equation}
W(\tilde{p}_k,{\langle s,d \rangle })
\equiv  W_s(\tilde{p}_k,d)=r_k\,,\ s\neq\Lambda\,.
\end{equation}
This means, by definition of Kolmogorov complexity, that
$K_W(r_k|s,d)=\min_{i\in\tilde{\set{K}}}|\tilde{p}_i|$, $s\neq\Lambda$.
Similarly from Eq.~(\ref{rs}), 
we have $K_U(r_k,s_k|d)=\min_{i\in{\set{K}}}|p_i|$
and therefore Eq.~(\ref{ShortestLengths}) becomes
\begin{equation} \label{AlmostThere}
K_W(r_k|s,d)=K_U(r_k,s_k|d)-K_U(s|d)+\kappa\,.
\end{equation}
Because $W(p,\langle\Lambda,d\rangle)=U(p,d)$ we have, for instance,
$K_U(s|d)=K_W(s,\langle\Lambda,d\rangle)$. Using this observation
to transform both terms at the right hand side of Eq.~(\ref{AlmostThere}), and 
choosing $s=s_k$ we have Eq.~(\ref{KwKu}) as required. $\Box$

Note that, since $U$ is an arbitrary prefix computer,
the above analysis provides a grouping of all
possible reference computers into naturally restricted classes.

\section{Acknowledgments}

It is my pleasure to acknowledge many helpful suggestions by
Jens G. Jensen, A.S. Johnson and Yuri Kalnishkan.

\end{document}